\documentclass[letterpaper]{article} 
\usepackage{aaai2026}  
\usepackage{times}  
\usepackage{helvet}  
\usepackage{courier}  
\usepackage[hyphens]{url}  
\usepackage{graphicx} 
\usepackage{natbib}  
\usepackage{caption} 
\usepackage{algorithm}
\usepackage{algpseudocode}
\usepackage{amsmath}

\usepackage{newfloat}
\usepackage{listings}
\DeclareCaptionStyle{ruled}{labelfont=normalfont,labelsep=colon,strut=off} 
\floatstyle{ruled}
\newfloat{listing}{tb}{lst}{}
\floatname{listing}{Listing}
\title{Towards Trustworthy Multi-Turn LLM Agents via Behavioral Guidance}
\author{Gonca G\"{u}rsun}
\affiliations{Bosch Center for Artificial Intelligence \\
Renningen, Germany}

\begin{document}

\maketitle

\begin{abstract}
Large Language Models demonstrate strong reasoning and generation abilities, yet their behavior in multi-turn tasks often lacks reliability and verifiability.  
We present a task completion framework that enables LLM-based agents to act under explicit behavioral guidance in environments described by reinforcement learning formalisms with defined observation, action, and reward signals. 

The framework integrates three components: a lightweight \textit{task profiler} that selects reasoning and generation strategies, a \textit{reasoning module} that learns verifiable observation–action mappings, and a \textit{generation module} that enforces constraint-compliant outputs through validation or deterministic synthesis.  
We show that as the agent interacts with the environment, these components co-evolve, yielding trustworthy behavior.
\end{abstract}


\section{Introduction}
\label{sec:intro}
Most real-world tasks from troubleshooting software to planning multi-step operations or interacting with users require agents to perform consistent action selection across turns and maintain constraint-compliant behavior generation throughout execution. Although recent advances~\cite{yao2023react, shinn2023reflexion, schick2023toolformer, autogpt2023, xu2024memgpt, wang2023voyager, babyagi2023} in agentic LLMs have improved their task completion abilities through mechanisms such as memory, tool use, and reflection, these mechanisms remain largely implicit and difficult to guide or steer, making it challenging for those building agentic systems on top of these models to maintain verifiable and reliable task completion~\cite{ganguly2025grammars, matton2025walktalk}.

Our goal is to provide LLM-based agents with a task completion framework that allows them to operate under explicit behavioral guidance.  
In this framework, \textit{trust} denotes the agent’s capacity to act in ways that are both \textit{verifiable} (its reasoning of action selection can be inspected and validated) and \textit{reliable} (its generated behaviors consistently comply with task constraints and environment feedback)~\cite{li2025certifiedtrustworthiness, decerqueira2025mappingtrustworthiness}.

\begin{figure}[!t]
    \centering
    \includegraphics[width=0.94\columnwidth]{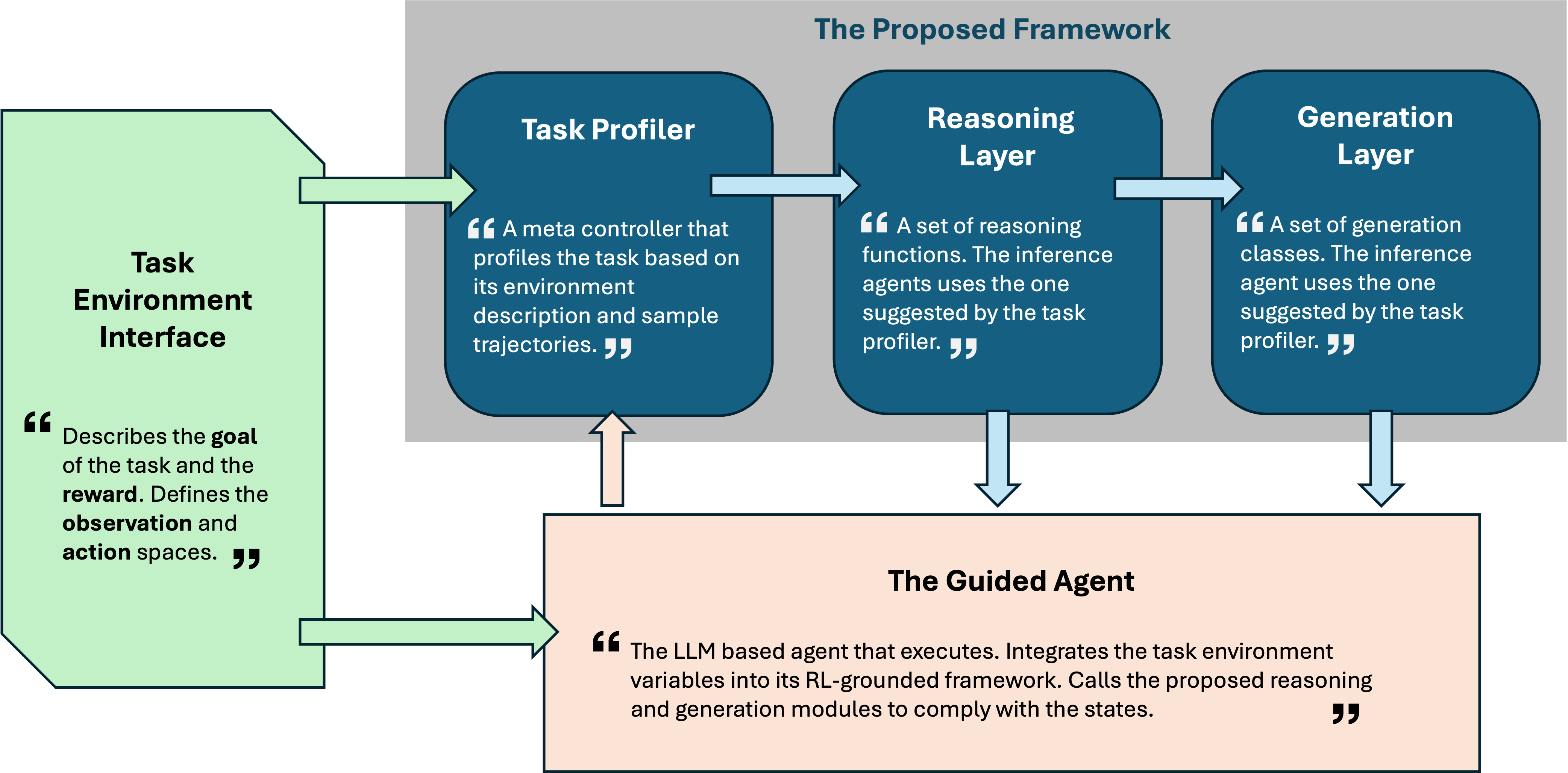}
    \caption{The Proposed Framework}
    \label{fig:framework}
\end{figure}

We target tasks described in reinforcement learning (RL) formalisms, where environments define actions, observations, rewards.
Within this setting, we develop a \textit{task completion framework} that enables LLM agents to learn to act in a verifiable and reliable manner. Our framework has three main components as shown in Figure~\ref{fig:framework} and described below.  

The first component is a lightweight \textit{task profiler} that analyzes the given task environment variables.
The profiler acts as a meta-learner, determines the task’s structural properties (e.g., temporal dependencies or constraint intensity), and guides the LLM agent toward the most suitable strategies for action selection and behavior generation.  

Building on this guidance, the second component, the \textit{reasoning module}, governs the selection of structured actions across temporal windows.  
It analyzes past trajectories from the agent’s task executions and extracts \textbf{observation–action mappings} that consistently yield high rewards.  
Guided by the task profiler, the reasoning module can adapt its temporal scope: in tasks where success depends on short-term decisions, it focuses on single-turn mappings, whereas in temporally dependent tasks, it aggregates information over longer horizons.  
The extracted mappings are stored as a persistent \textit{procedural memory} and integrated with the underlying LLM’s native reasoning during subsequent task executions.  
Throughout the paper, we refer to these mappings as \textit{rules} and use two terms interchangeably\footnote{Note that we use the term \textit{rules} broadly to include both explicit conditional patterns and higher-level strategies.}.

Finally, the third component, the \textit{generation module}, ensures constraint-compliant behavior generation by validating or revising the agent’s outputs so that they satisfy all task constraints and reasoning-derived mappings.  
Its compliance strategy is determined by the task profiler: for lightly constrained tasks, the module may simply verify the validity of the model’s native output, whereas for constraint-heavy tasks, it employs structured procedures such as deterministic enumeration or online code generation.  
In these cases, the module uses environment variables and reasoning mappings as input specifications to generate valid, verifiable outputs that align with task feedback.

As the agent executes a given task, these components interact continuously:  
the task profiler refines its understanding of the environment over epochs\footnote{An \textit{epoch} denotes a complete cycle of multiple trajectories, each \textit{trajectory} being a full sequence of observation–action–reward steps.}.
,  
the reasoning module progressively learns better observation–action mappings from collected trajectories.
,
and the generation module evolves along it, adapting its output strategies to the updated reasoning state.  
Together, they steer the native behavior of the underlying LLM into a transparent, feedback-guided process where every action is both verifiable and reliable.

In this paper, we present the first evaluation results of the proposed framework on two representative multi-turn environments: \textit{Guess My Number} and \textit{Wordle}.  
The evaluation focuses on three complementary metrics: \textbf{a) average task completion reward}, \textbf{b) consistency of action selection}, and \textbf{c) compliance with} constraints, which together capture agent verifiability and reliability.  
Across both environments, agents guided in our framework consistently outperform native baselines with and without in-context learning.

Figure~\ref{fig:framework} shows the high-level interaction of our framework with a given task's environment interface and the LLM agent assigned for the task. Our framework provides a generic \emph{task environment interface} to describe the execution environment (i.e. observation and action state variables, reward mechanism, and goal description) of a given task. Once the environment is described, all components of the framework and the LLM agent conform to this interface. 

In the following, we first describe the rest of the components in Figure~\ref{fig:framework} in detail. Then we present
how they work together in Algorithm~\ref{alg:framework_loop}. Finally, in the experiments section, we exemplify their functions through two sample tasks.

\subsection{The Agent with RL Prompting Framework}
\label{sec:rl_prompting}
\begin{figure}[htbp]
    \centering
    \includegraphics[width=0.94\columnwidth]{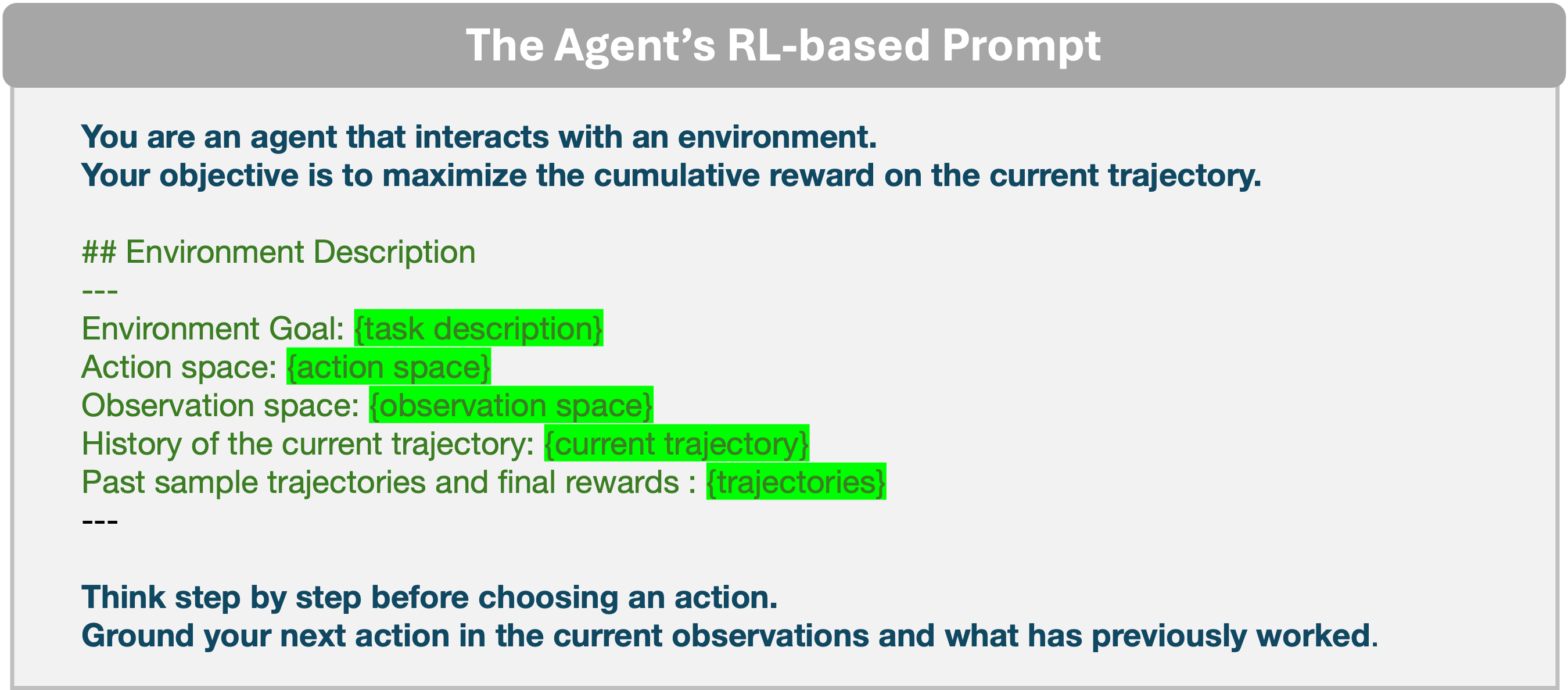}
    \caption{The main execution prompt of the LLM agent. Placeholders in \{\} are populated during execution.}
    \label{fig:baseline-prompt-template}
\end{figure}

At the foundation of our framework lies an \textbf{action--observation--reward interaction loop} grounded in reinforcement learning. 
This prompting backbone (see Figure~\ref{fig:baseline-prompt-template}) supplies a structure for the LLM’s native generation and sequential task execution. By embedding the model in an explicit loop of actions, observations, and rewards, the backbone enforces temporal coherence and prevents the agent from treating each turn as an isolated completion. 

Concretely, the prompting backbone interfaces with a task environment through a standardized schema exposing three fields: \emph{observations}, \emph{actions}, and \emph{rewards}. 
In addition, at each turn, the LLM agent receives two kinds of structured context:
(1) an explicit record of the \textbf{current trajectory history} (all past observations and actions in the ongoing run), anchoring subsequent \emph{action selection} in accumulated progress; and
(2) optionally, a set of \textbf{past trajectories with final rewards} to simulate in-context learning (ICL) from prior experience.
While the ICL field is not required for the loop itself, it allows us to assess whether contextual exposure to past successes improves the task completion or not.

When used on its own, i.e. without our proposed framework this backbone yields a \textbf{prompting-only baseline agent}. 
It provides the minimal structure for sequential \emph{action selection} and produces an auditable interaction trace, allowing us to evaluate the intrinsic behavior of LLMs under feedback-guided prompting. 
In the next sections, we augment this backbone with a task profiler that supplies \emph{behavioral guidance}, reasoning, and generation modules.

\subsection{Task Profiler}
\label{sec:task_profiler}
One of the main challenges in multi-turn environments is that different tasks demand different styles of behavioral guidance at different temporal stages of the task execution.  
Some tasks require rapid, turn-local responses to new feedback, while others rely on long-horizon bookkeeping or strict enforcement of cumulative constraints.  
Without a mechanism to detect these structural differences, LLM agents tend to drift between inconsistent reasoning modes, reducing reliability over extended interactions~\cite{guerdan2025llmjudge}.

The \textit{task profiler} provides this adaptive guidance layer.  
It analyzes the task environment and identify key features of the task and informs how the agent should adjust its reasoning and generation strategy to increase its task completion success.
Recent brain-inspired modular architectures similarly highlight the value of separating task analysis, planning, and constraint monitoring as coordinated processes to improve multi-step reliability~\cite{webb2025map}.

The profiler does not solve the task directly; instead, it determines \textbf{how the desired behavior should be generated} for successful task completion. In that regard, the task profiler acts similar to a meta-level learner. 

In this paper, we make our first attempt to design a task profiler grounded on principles from cognitive science~\cite{Newell_1990}, symbolic AI~\cite{Fikes_1971, SolarLezama_2008}, and reinforcement learning~\cite{Sutton_1999, Garcia_2015} that emphasize adaptive, interpretable behavior.  
Following Newell’s concept of \emph{flexible human learning}~\cite{Newell_1990}, it is designed to identify the structural demands of a task and select reasoning and generation strategies suited to its temporal and constraint patterns.

\begin{figure*}[htbp]
    \centering
    \includegraphics[width=0.72\textwidth]{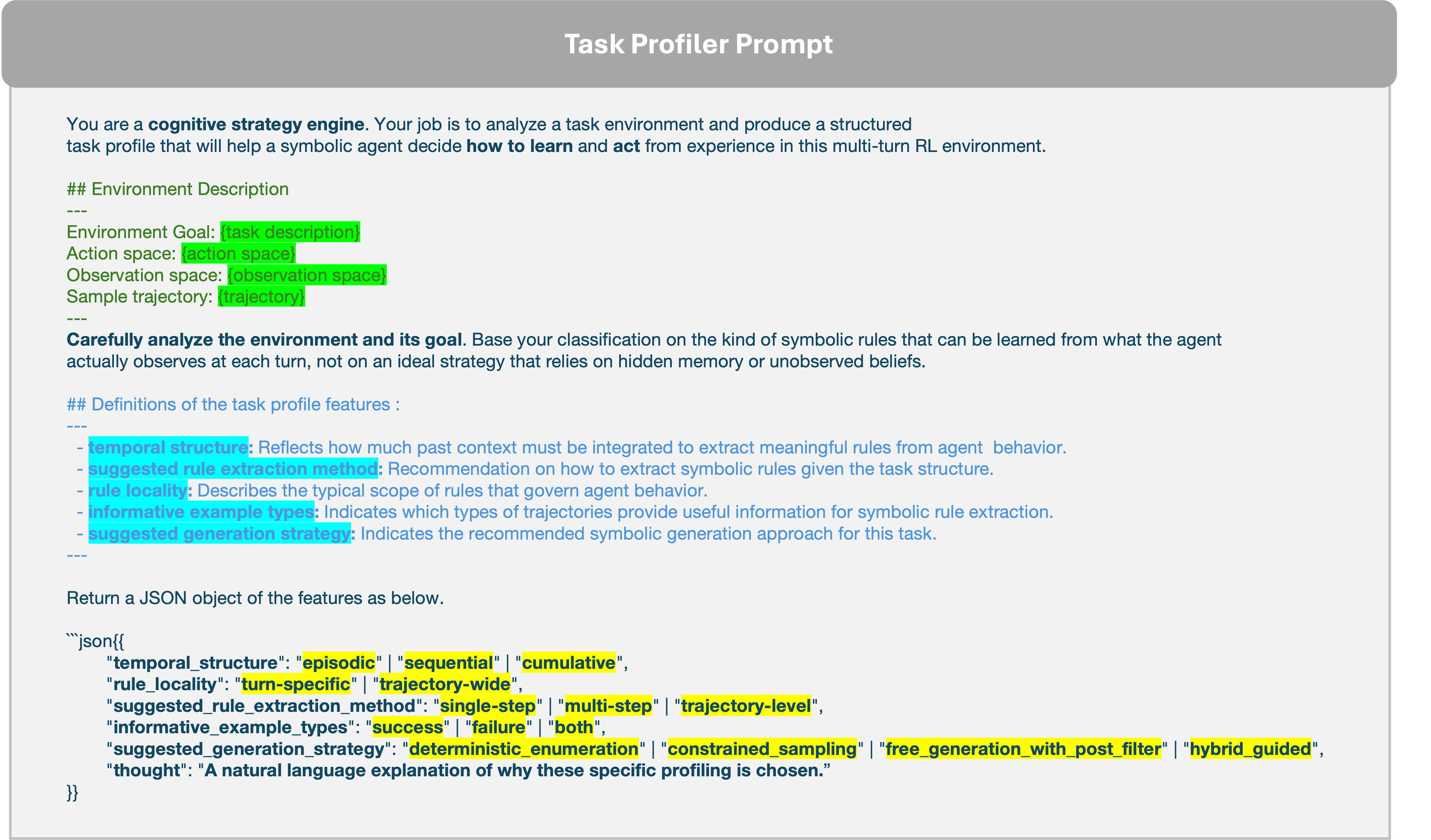}
    \caption{Prompt design for the LLM-based task profiler.}
    \label{fig:task_profiler_prompt}
\end{figure*}

Our current implementation represents a minimal instantiation of such a profiler. We implement the profiler as a lightweight, LLM-based function prompted to act as a \textit{cognitive strategy engine} (see Figure~\ref{fig:task_profiler_prompt}). The task of this engine is to map a given environment to a compact set of \textit{task features} such as temporal structure, suitable generation strategy etc. as defined in the prompt to capture the essential behavioral dimensions of a task. Later, it can be extended toward more nuanced or data-driven profiling strategies that incorporate richer environment cues.

By adaptively identifying the appropriate behavioral regime for each task, the profiler allows guided agents to sustain verifiable reasoning and reliable task execution across structurally diverse environments.

\subsection{Reasoning Layer}
\label{sec:reasoning}
The reasoning layer gives the agent a persistent, interpretable memory of what has led to successful task completion in the past.  
Instead of treating every step as a new generation, it analyzes past trajectories and learns \textit{reusable action rules} that connect what the agent \textit{observes} with what it \textit{should do next} to achieve reward.  
These rules accumulate over time, forming a growing library of ``if condition $\rightarrow$ action'' mappings that guide behavior across multiple turns.  
This builds an additional layer of reasoning on top of the underlying LLM’s native capabilities, allowing the agent to maintain interpretable logic that persists beyond the model’s immediate token-by-token reasoning.

The reasoning layer adapts its operation based on the \textit{task profile} provided by the profiler.  
If the profiler classifies a task as \textit{sequential}, the layer focuses on analyzing how the environment evolves with each subsequent turn and identifying which actions consistently lead to successful task completion.  
If the task is \textit{cumulative}, it integrates information across turns, maintaining long-horizon eliminations and constraints that shape later choices.  
This adaptivity ensures verifiable and consistent behavior aligned with the structural demands of each task.

Operationally, the reasoning layer is implemented as a collection of LLM-based functions, each aligned 
with a specific reasoning strategy prescribed by the task profile.  
These functions consume task environment variables and past trajectories.

The reasoning layer examines past trajectories at a temporal granularity suggested by the profiler and then identifies regularities that consistently led to reward. 
For example, in a task where the profiler suggests a \textit{one-step temporal structure}, when a non-zero reward is observed at turn $t\!+\!1$, the module inspects the previous turn $t$: the observation at $t$, the action taken, and the resulting outcome.  
It then asks the LLM to express this relationship as a rule of the form:

\begin{center}
\texttt{if [observation condition at turn $t$], then the best action is [action at $t\!+\!1$]}.
\end{center}

Each discovered rule is stored in a structured format inside a central \textit{Rule Bank}, along with its success rate and usage history \footnote{Due to lack of space, we omit the details of the Rule Bank and how we manage (register, test, filter) rules over subsequent runs (trajectories) of the same task. The further details can be found in the code base.}. 
Over time, this collection becomes an explicit, auditable representation of the agent’s learned behavioral logic, tested across many runs of the same task.  
When a familiar condition reappears, the corresponding rule can be applied immediately, providing a verifiable and efficient way to select actions. Importantly, the reasoning layer does not execute actions directly. It serves as a stable substrate of verified logic that constrains the generation layer.

\subsection{Generation Layer}
\label{sec:generation}
We separate reasoning and generation in our framework because they are conceptually and functionally distinct cognitive processes. Splitting them allows for greater interpretability and reliability. In our framework, the reasoning layer focuses on understanding and structuring the task logic, while the generation layer is responsible for executing that logic as valid, reliable actions.

While reasoning keeps track of how observations relate to valid actions, generation executes this knowledge, i.e. it ensures that outputs remain valid under all environment constraints and consistent with verified reasoning.  
This is where the agent’s planned behavior becomes visible: every generated action must pass through a validation step before being accepted.

The generation layer is implemented as a class of tools. The task profiler prescribes the most suitable one among them depending on the task’s structural complexity.  
For lightly constrained tasks, the profiler may recommend a direct generation mode, where the model’s native output is simply checked for validity against the reasoning state.  
For tasks that are cumulative or constraint-heavy, the profiler prescribes structured generation procedure (e.g. deterministic enumeration or guided sampling) to ensure that all generated actions satisfy accumulated task rules.  
Typical examples include structured games such as Sudoku or Wordle, where output validity is tightly constrained by the evolving game state. In these environments, the profiler classifies the problem as cumulative and constraint-heavy and recommends deterministic enumeration implemented as code over the candidate set.

At each turn, the reasoning layer proposes a structured action containing both the intended output and its associated set of constraints.  
Before committing this output, the generation layer validates it using the environment’s built-in validity check.  
If the proposed action violates any active constraint, e.g. repeating an invalid candidate or breaking positional rules) the system automatically falls back to deterministically enumerating all valid candidates. This fallback filters the candidate set for constraint compliance and selects the first valid option (or a random one, if permitted). Through this mechanism, every output remains \textbf{verifiably valid} with respect to both environment feedback and reasoning guidance.

Having introduced each module independently, we now describe how they operate together in Algorithm~\ref{alg:framework_loop}.

\begin{algorithm}[!t]
\caption{Task Execution Loop for the Guided Agent}
\label{alg:framework_loop}
\small
\textbf{Definitions:} 
$A$: action space — set of valid actions. \\
$a_t\!\in\!O$: action taken by the agent at turn $t$. \\
$O$: observation space — set of observations. \\
$o_t\!\in\!O$: observed state at turn $t$. \\
$r_t\!\in\!\mathbf{R}$: reward value received after executing $a_{t-1}$. \\
$f_{\text{reward}}:(o_t,a_t,o_{t+1},h_t)\!\mapsto\!\mathbf{R}$ — scalar reward function. \\
$f_{\text{validity}}:(a_t,o_t,h_t,\mathcal{E})\!\mapsto\!\{\text{True},\text{False}\}$ — constraint/format check. \\
$f_{\text{step}}:(a_t)\!\mapsto\!(o_{t+1},r_{t+1},d_{t+1})$ — environment transition. \\
$f_{\text{reset}}:()\!\mapsto\!o_0$ — resets environment. \\
$g_{\text{done}}:(o_{t+1},r_{t+1},h_{t+1})\!\mapsto\!\{\text{True},\text{False}\}$ — task completion predicate.\\
$\mathcal{E}=\{A,O,f_{\text{reward}},f_{\text{validity}},f_{\text{step}},f_{\text{reset}},g_{\text{done}}\}$ - task environment interface. 

\begin{algorithmic}[1]
\Require Task environment $\mathcal{E}$
\State Initialize empty RuleBank $\mathcal{R}\leftarrow\emptyset$
\State Initialize task profile $\mathcal{P}\leftarrow$ None
\For{epoch $e=1$ to $E$}
    \If{$e==k$} \Comment{after initial warm-up}
        \State $\mathcal{P}\leftarrow$\textbf{TaskProfiler}($\mathcal{E}$)
        \State Select reasoning function $f_{\text{reason}}$, generation operator $f_{\text{gen}}$ from $\mathcal{P}$
    \EndIf
    \For{trajectory $\tau=1$ to $T$}
        \State $o_0\leftarrow f_{\text{reset}}()$, $h\leftarrow\emptyset$
        \While{not $g_{\text{done}}$}
            \State Compose LLM prompt using $\mathcal{E}$, $h$, and applicable rules $\mathcal{R}$
            \State $a_t\leftarrow$\textbf{LLM.generate}($f_{\text{reason}},h$)
            \State Valid$\leftarrow$\textbf{ValidityCheck}($a_t,h,\mathcal{E}$)
            \If{not Valid}
                \State $a_t\leftarrow$\textbf{FallbackGenerate}($f_{\text{gen}},\mathcal{E}$)
            \EndIf
            \State $(o_{t+1},r_{t+1},d_{t+1})\leftarrow f_{\text{step}}(a_t)$
            \State Append $(o_t,a_t,r_{t+1})$ to trajectory history $h$
            \State $t\leftarrow t+1$
        \EndWhile
        \State Store trajectory $(h,r_{t+1})$ in epoch log
    \EndFor
    \State \textbf{ReasoningUpdate}($\mathcal{R}$, epoch log)
    \State (Optional) $\mathcal{P}\leftarrow$\textbf{TaskProfiler} if task dynamics shift
\EndFor 
\end{algorithmic}
\end{algorithm}

\section{Experiments}
\label{sec:experiments}

\subsection{Tasks}
\label{sec:experiment_tasks}
\begin{figure}[htbp]
    \centering
    \includegraphics[width=0.94\columnwidth]{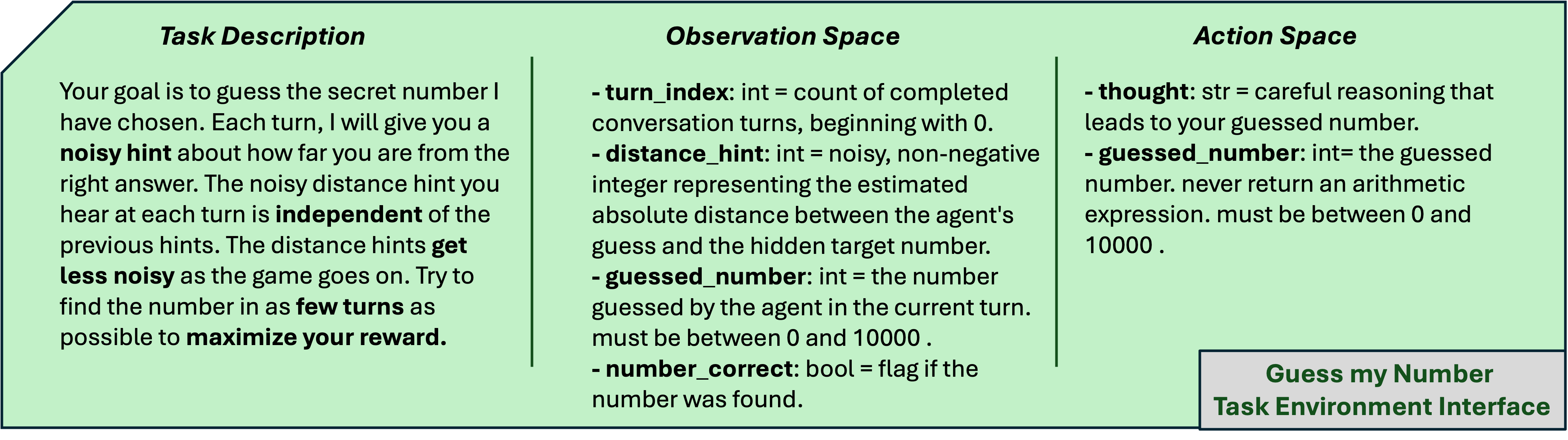}
    \caption{Guess My Number (GmN) environment interface.}
    \label{fig:gmn-task-environment}
\end{figure}

\begin{figure}[htbp]
    \centering
    \includegraphics[width=0.94\columnwidth]{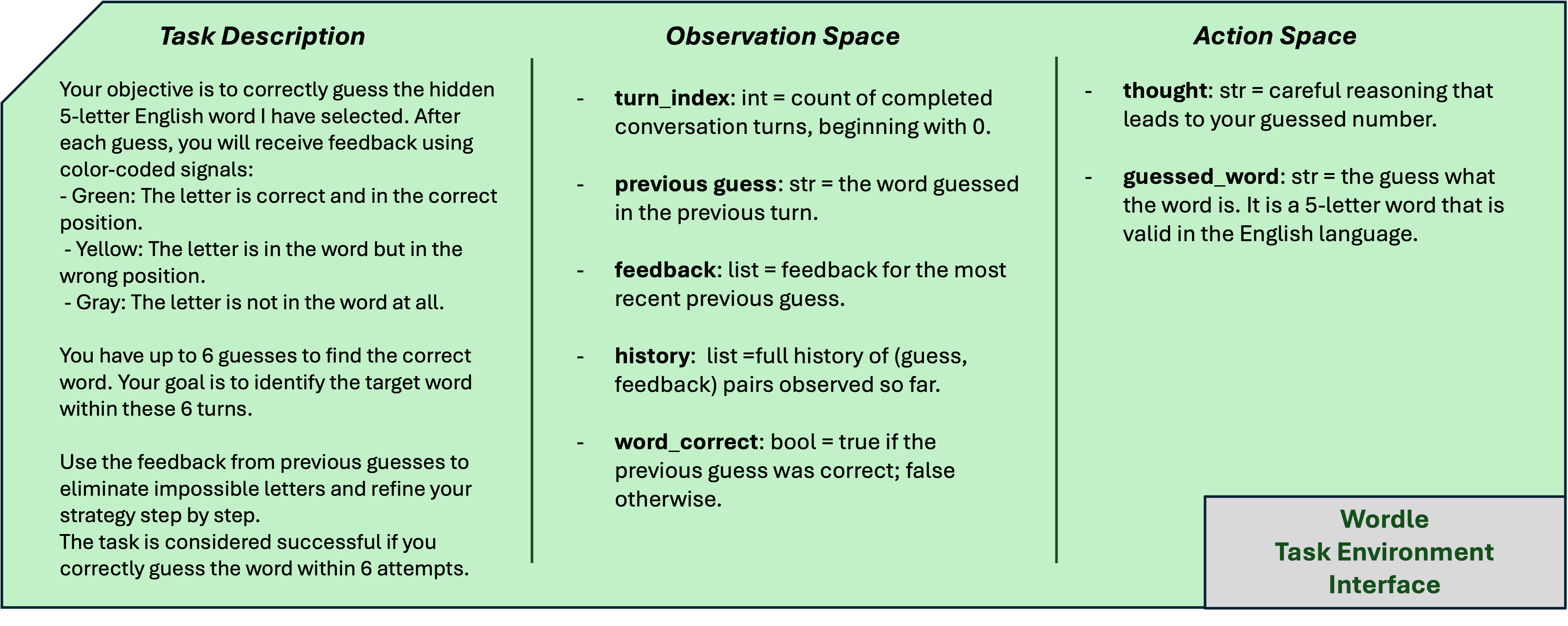}
    \caption{Wordle environment interface.}
    \label{fig:wordle-task-environment}
\end{figure}

For evaluation, we select two intuitive yet structurally distinct 
multi-turn game tasks: \textit{Guess My Number (GmN)} and \textit{Wordle}. For each game, we implement their task environments as shown in Figure~\ref{fig:gmn-task-environment} and Figure~\ref{fig:wordle-task-environment}, respectively. 

\paragraph{Guess My Number (GmN).}
At the start of each trajectory (one complete run of the game), a secret target number is sampled within the range $[0, 10000]$. The task of the LLM agent is to correctly guess this number. At each turn, the agent proposes a guess and receives a noisy distance hint indicating how far the guess is from the target. Each trajectory lasts for at most 15 turns. If the agent identifies the correct number on turn $t$, it receives a reward of $100 / t$; otherwise, the reward is $0$.

The distance hint noise is generated by $\text{noise} = 1000 \cdot 0.2^{t}$, where $t$ is the current turn index. This generating function is hidden from the agent, but the task description specifies that (i) the noise decreases as $t$ increases, and (ii) the noise at each turn is independent of past turns. After $t > 5$, the noise becomes negligible, meaning the hint provides the exact distance to the secret number. These temporal regularities are the key to successful action selection and higher overall reward. We evaluate whether the agent learns to recognize and exploit these patterns, verifying them through reasoning and reusing them across epochs to guide future guesses.

\paragraph{Wordle.}
At the start of each trajectory (one complete run of the game), a secret five-letter English word is sampled. 
The task of the agent is to correctly guess this word. At each turn, the agent proposes a candidate 
and receives letter-level feedback: a letter is marked as \textit{correct} if it is in the right position, \textit{misplaced} if it appears elsewhere, and \textit{absent} if it does not occur in the target. Each trajectory lasts for at most 6 turns. 
If the agent guesses the correct word at any turn $t$, it receives a reward of $100$; otherwise, the reward is $0$.

Wordle requires the agent to manage a set of cumulative hard constraints that evolve across turns.  
Each feedback update specifies which letters and positions are fixed, which letters are excluded, and which must appear elsewhere.  
To succeed, unlike in GmN, the agent must maintain a trajectory-wide record of these constraints and ensure that every new guess complies with all accumulated information.  
We evaluate whether the agent learns to track these evolving constraints and generate reliable outputs that remain valid and consistent across turns.

\subsection{Setup}
\label{sec:experiment_setup}
\paragraph{Agent Variants.}
We evaluate two categories of agents:  
a \textbf{baseline agent} that operates without our framework, and a \textbf{guided agent} that uses the same prompting backbone, but runs within our task completion framework.

The \textbf{baseline agent} relies only on the RL prompting backbone, interacts with the environment through the standard observation–action–reward interface.  
At each turn, the LLM generates an action directly from the prompt context, without any persistent reasoning state or additional constraint check.  
The baseline agent is tested in two variants: with and without in-context learning.
In the in-context variant, a small set of randomly sampled successful trajectories is included in the prompt (see Figure~\ref{fig:baseline-prompt-template}) at the start of each epoch.  The \textbf{guided agent} augments the same backbone with our proposed framework. 

\paragraph{The Runs \& Model}
All agents are evaluated for 30 epochs, with 20 trajectories each in both tasks. Performance metrics are reported with 95\% confidence intervals. 
In all experiments, we use GPT-4.1-mini, a non-reasoning model, as the underlying LLM.  
This choice is intentional: it allows us to isolate the architectural contributions of our proposed framework 
from factors related to model scale or intrinsic reasoning ability. Nonetheless, the framework itself is model-agnostic, and future work will evaluate its effectiveness 
with larger, reasoning-capable agentic LLMs.

\subsection{Results}
\label{sec:results}

\begin{figure}[htbp]
    \centering
    \includegraphics[width=0.96\columnwidth]{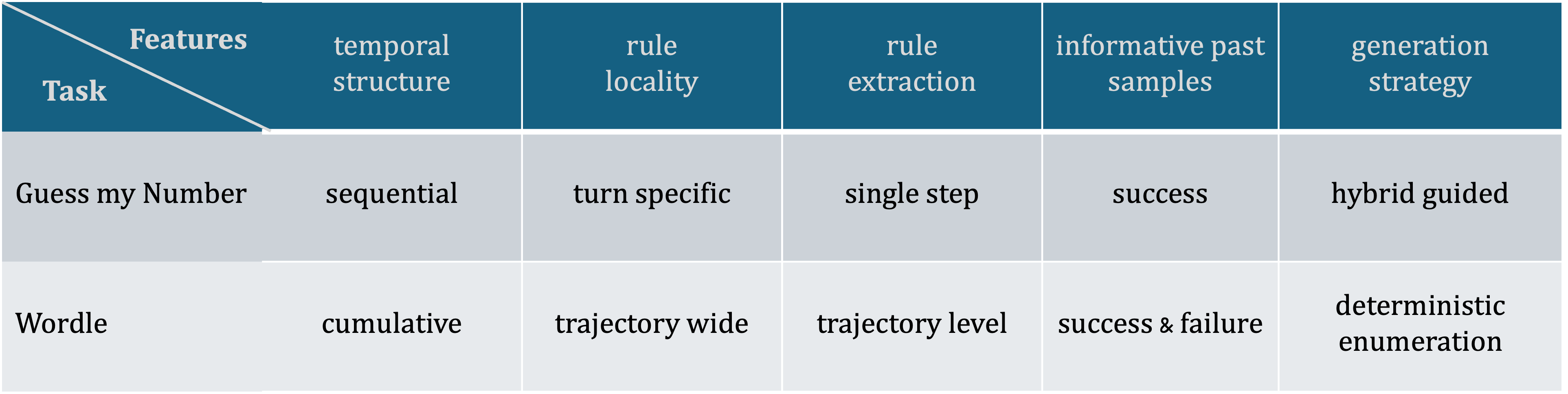}
    \caption{Task profiler generated outputs.}
    \label{fig:task_profiler_outputs}
\end{figure}

To start with, Figure~\ref{fig:task_profiler_outputs} presents the output 
of the task profiler for GmN and Wordle for the existing task structure categories.
Throughout the experiments below, the \textit{guided agent}
behaves according to these strategies. The task profiler is run at the end of
each epoch and each time consistently outputs the strategies as presented.

\subsubsection{GmN Evaluation}
\label{sec:gmn_evalaution}
\begin{figure}[htbp]
    \centering
    \includegraphics[width=0.94\columnwidth]{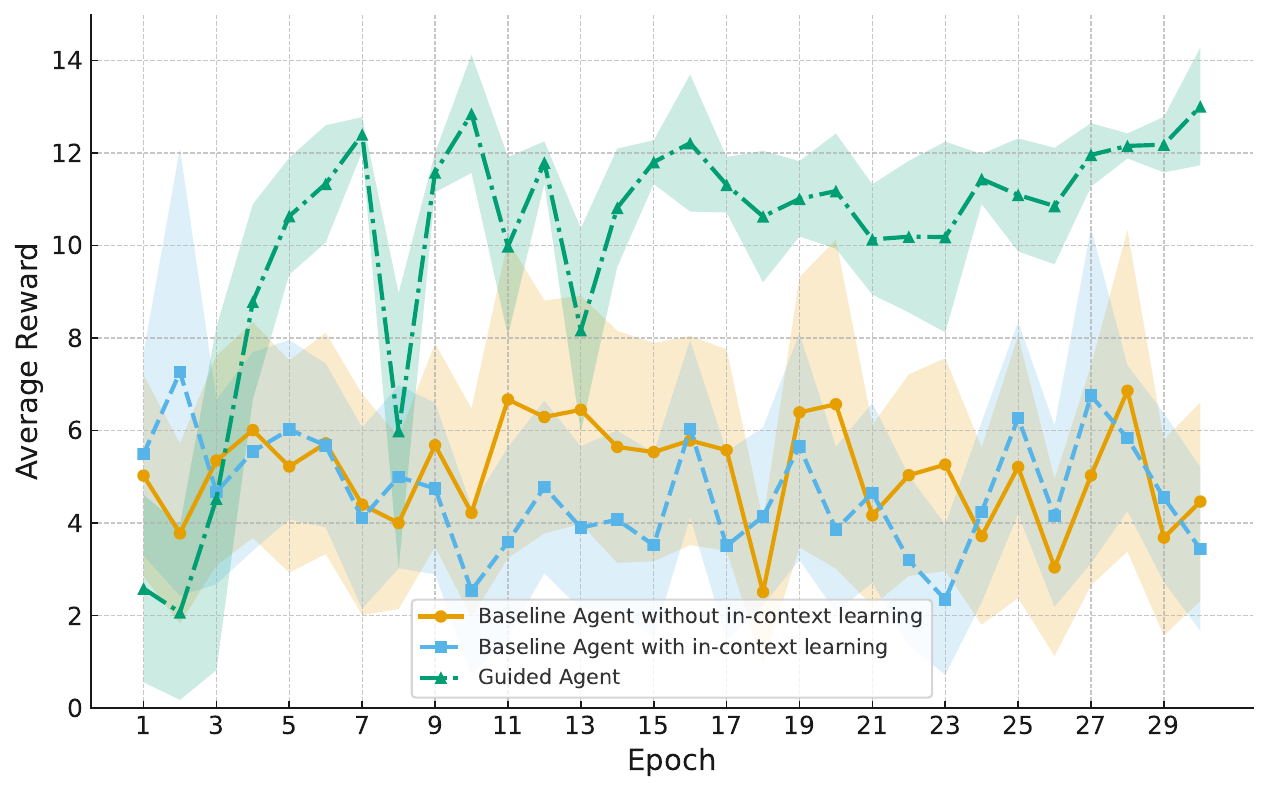}
    \caption{GmN: Average reward per epoch, each epoch is averaged over 20 trajectories, and presented with a 95\% CIs.}
    \label{fig:gmn_avg_reward}
\end{figure}

\paragraph{Task Success and Reasoning Consistency.}
Figure~\ref{fig:gmn_avg_reward} shows the average cumulative reward per epoch. 
The \textit{baseline agents}, with or without in-context learning, show no consistent improvement over time. This indicates that simple exposure to past trajectories is insufficient to achieve reliable behavior in a temporally structured task.  

In contrast, the \textit{guided agent} achieves steady improvement and stabilizes 
at a higher average reward. At the end of each epoch, our framework's reasoning module, guided by the profiler, analyzes the \textit{successful} trajectories focusing on what actions over the observed state yielded a correct guess. 

At first, the mappings (aka. rules) derived from these past trajectories tend to overfit to the actual values of the observed variables and do not generalize well, 
but as the epochs proceed the reasoning module
learns to generalize them by verifying their validity in the new trajectories. In fact, the practice of discovering new rules and testing them shows itself in the performance decline in epochs 8, 11, 13. This behavior corresponds to exploration vs. exploitation in reasoning until it converges to a high reward state. 

\begin{figure}[htbp]
    \centering
    \includegraphics[width=0.84\columnwidth]{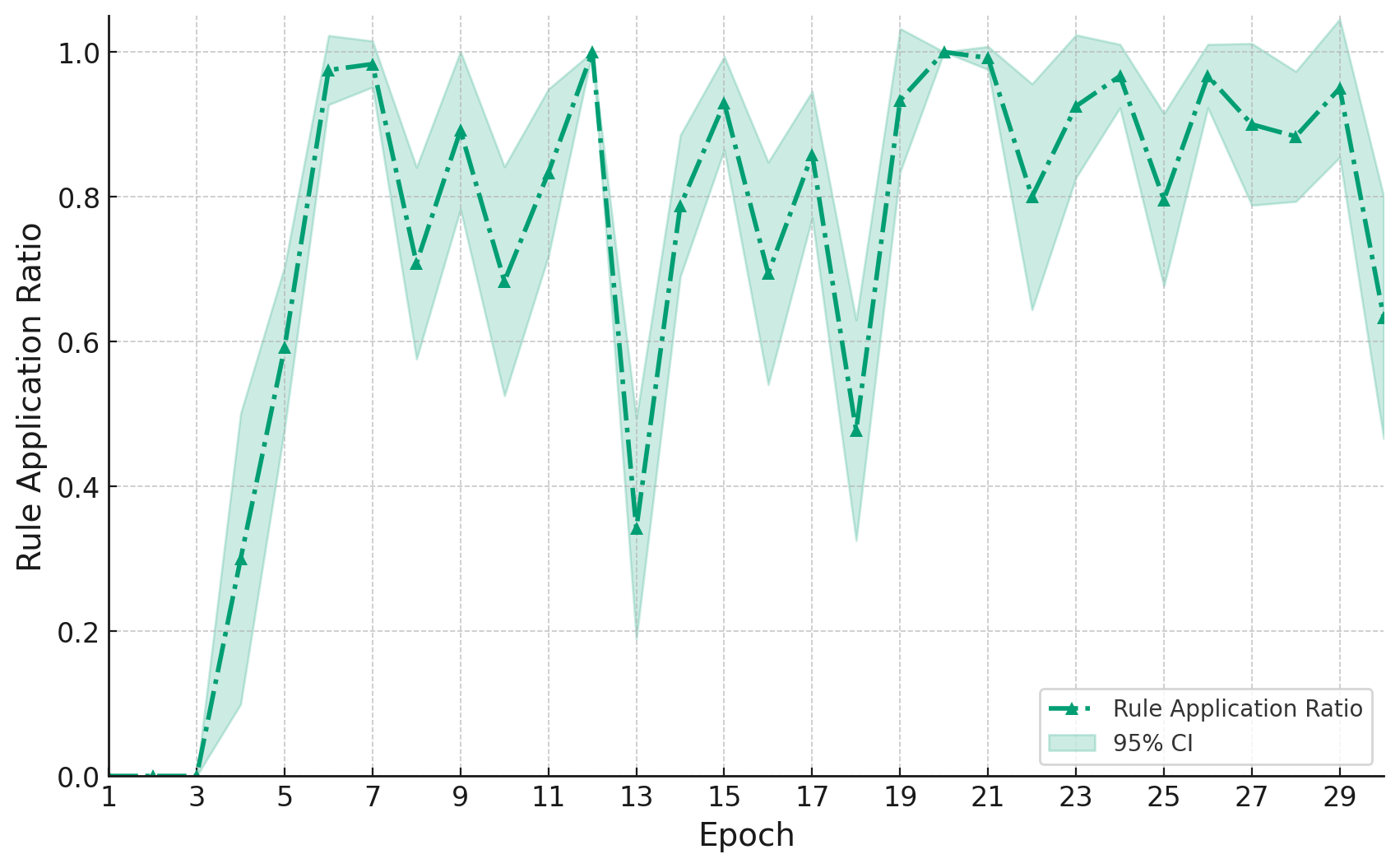}
    \caption{GmN: Proportion of turns extracted rules are correctly applied by the guided agent.}
    \label{fig:gmn_rule_application_ratio}
\end{figure}

After epoch 15, the rules stabilize, marking the transition from ad-hoc reasoning to generalized, consistent reasoning. 
We quantify this progression in the next section using the \textit{reasoning consistency ratio} (Figure~\ref{fig:gmn_rule_application_ratio}), which measures how reliably the agent applies these learned mappings across turns.
Overall, these rules serve as repeatable patterns and show that the agent transitions from ad-hoc reasoning to generalized, consistent reasoning over time.

\begin{figure}[htbp]
    \centering
    \includegraphics[width=0.96\columnwidth]{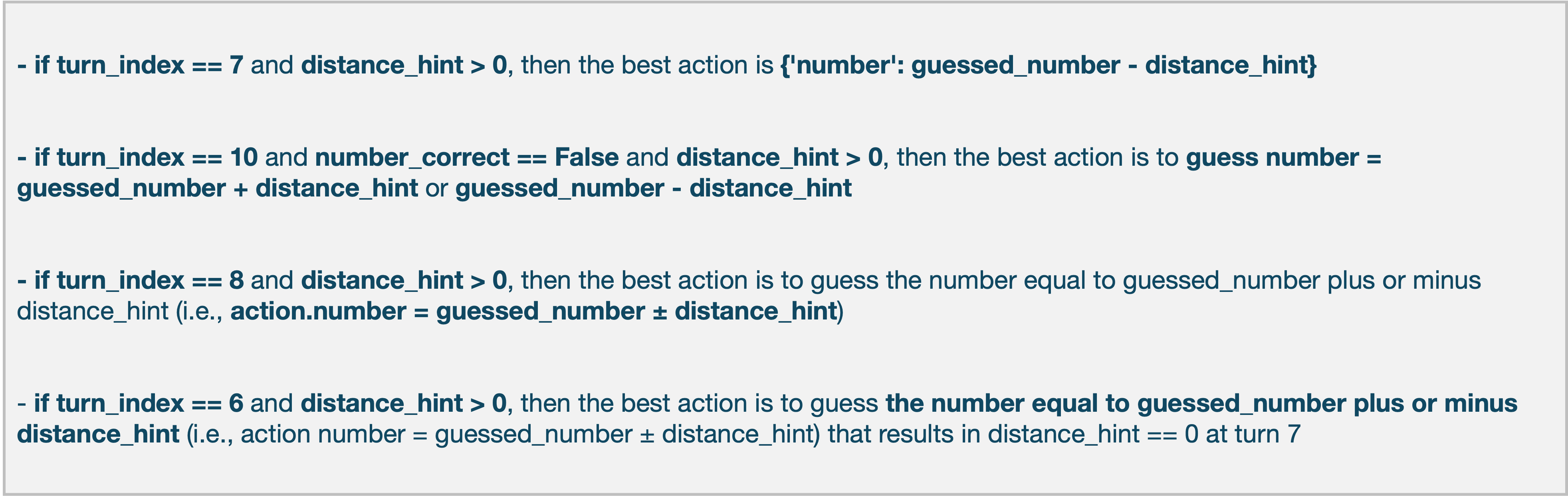}
    \caption{GmN: Verified rules after epoch 15}
    \label{fig:reasoning_rules_gmn}
\end{figure}

\paragraph{Verifiability and Reliability.}
We assess \textit{verifiability of reasoning} using the \textit{reasoning consistency ratio}:  
the fraction of turns in which an \emph{applicable rule} from the learned mapping set is correctly invoked for the current observation.  
Here, an applicable rule refers to a mapping whose preconditions match the features of the current observation state.  
As shown in Figure~\ref{fig:gmn_rule_application_ratio}, this ratio increases steadily across epochs, indicating that the agent’s reasoning becomes progressively more stable and rule-consistent over time.  
Occasional dips correspond to exploratory epochs in which newly discovered rules are tested or refined as explained above.

We assess \textit{reliability of behavior} via outcome variance:  
the guided agent exhibits narrower confidence intervals in average reward (Figure~\ref{fig:gmn_avg_reward}),  
signaling lower across-trajectory variability and more predictable performance.  
Together, these results show that our framework improves both \emph{verifiability} (reasoning that can be checked against explicit mappings) and \emph{reliability}(behavior that remains stable and consistent across runs.).

\subsubsection{Wordle Evaluation}
\label{sec:wordle_evalaution}
\begin{figure}[htbp]
    \centering
    \includegraphics[width=0.94\columnwidth]{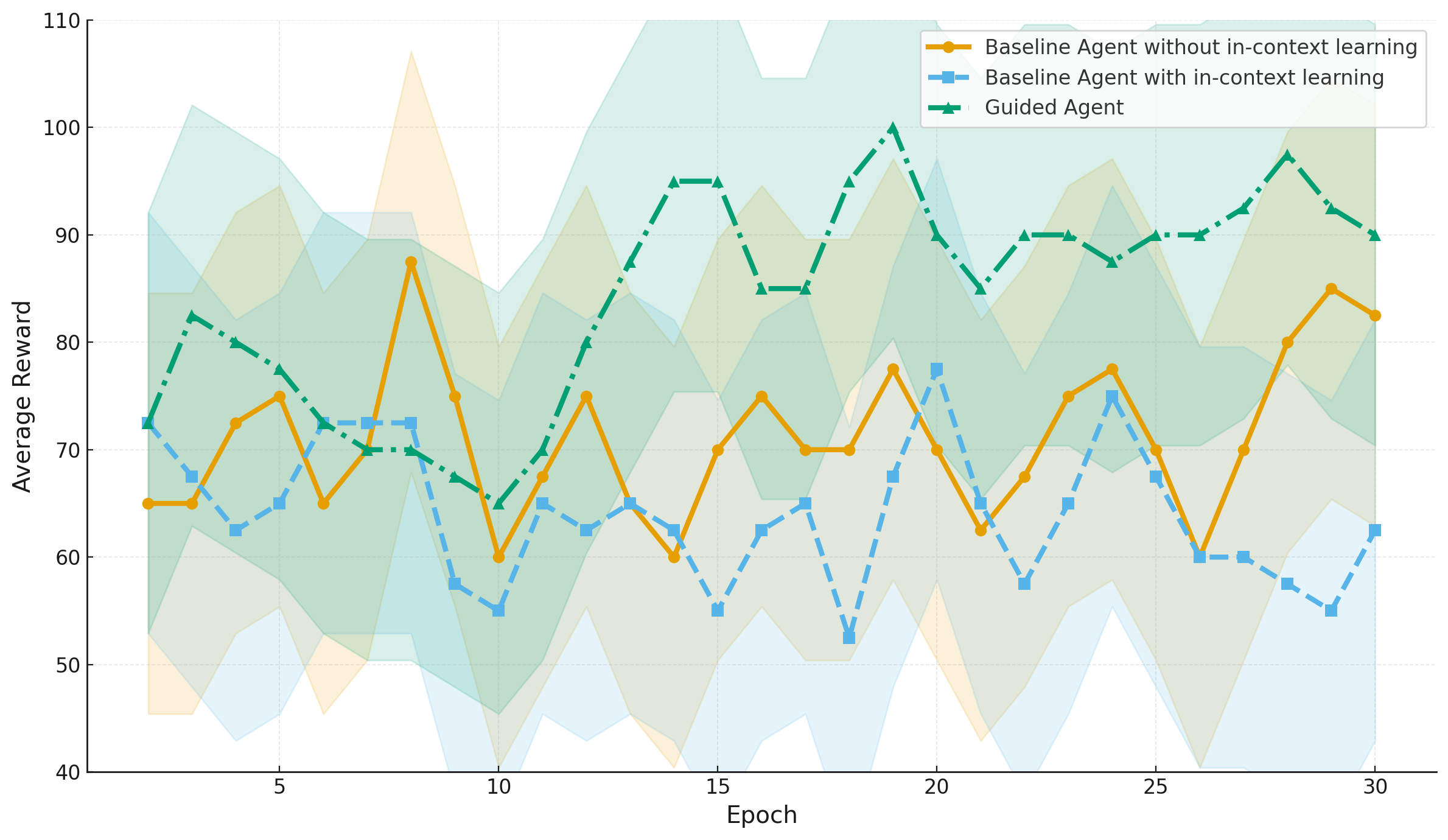}
    \caption{Wordle: Average reward per epoch, each epoch is averaged over 20 trajectories, and presented with a 95\% CIs.}
    \label{fig:wordle_avg_reward}
\end{figure}

\paragraph{Task Success and Constraint Compliance.}
Most of the performance gains in the Wordle task stem from the generation module. Therefore, to isolate the contribution of the generation module, we activate it only after epoch 10;  
during the first ten epochs, the guided agent operates without code-based output generation, relying solely on the reasoning module.  
This allows us to directly observe the effect of introducing programmatic constraint enforcement on behavioral stability.

Figure~\ref{fig:wordle_avg_reward} shows the average cumulative reward across epochs.  
Baseline agents show no consistent improvement: although they often restate constraints correctly in natural language, their generated outputs frequently violate them.  
Providing in-context demonstrations of successful trajectories yields no measurable benefit, confirming that contextual exposure alone does not enable consistent constraint enforcement.
 
The \textit{guided agent} eliminates constraint violations because the task profiler correctly
classifies the environment as \textit{cumulative and constraint-heavy}.
Based on this classification, the profiler guides the agent to use \textit{code-based} generation, where outputs are programmatically generated to satisfy all accumulated constraints.  

\paragraph{Verifiability and Reliability.}
We define \textit{verifiability of generation} as the agent’s ability to produce outputs whose correctness can be objectively checked against known task constraints.
We quantify this using the \textit{constraint-compliance ratio}, i.e. the proportion of turns in which generated outputs satisfy all active constraints defined by feedback so far.

\begin{figure}[t]
    \centering
    \includegraphics[width=0.94\columnwidth]{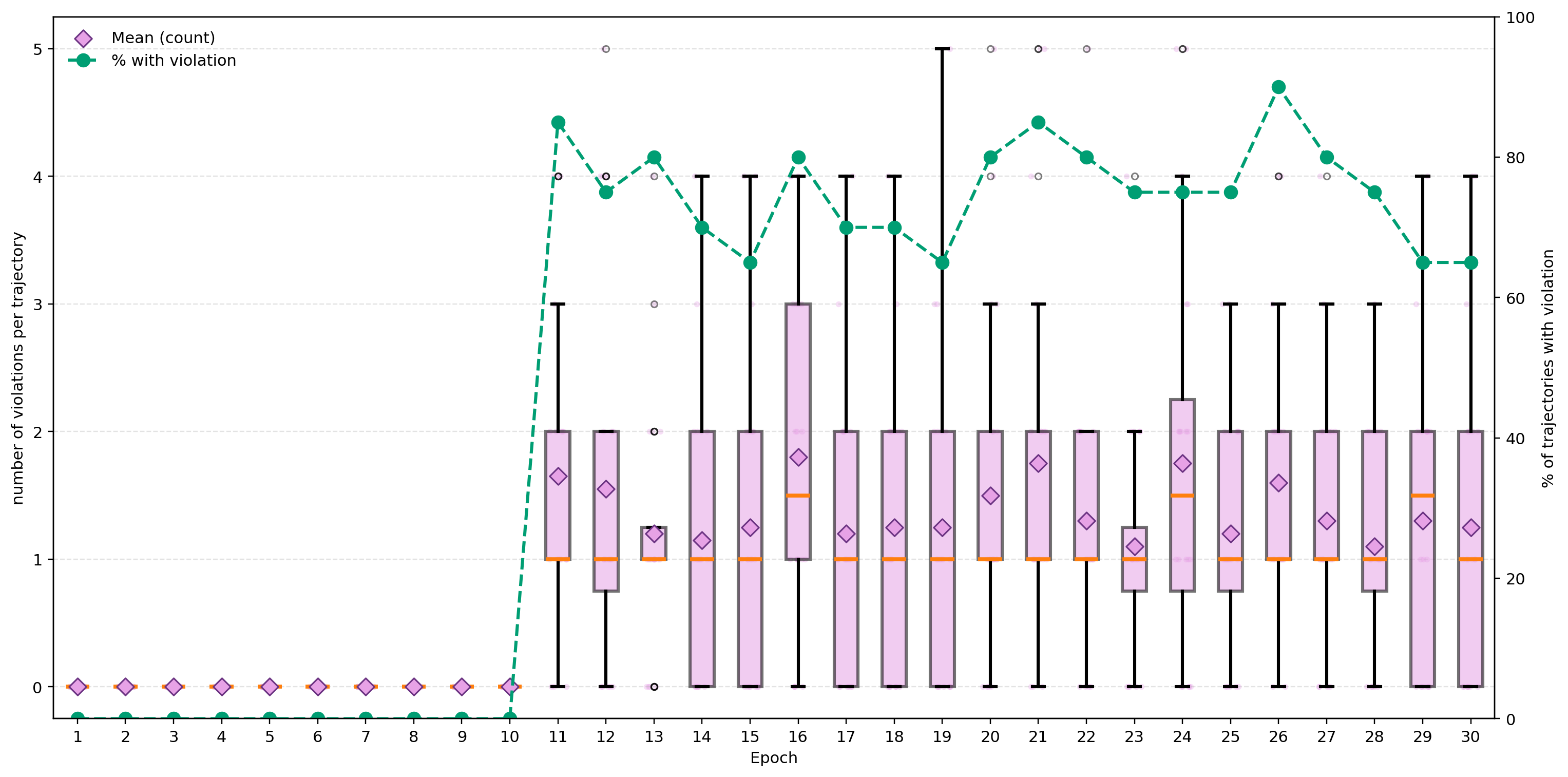}
    \caption{Wordle: Constraint-compliance and recovery rate of the guided agent. 
    Each box represents the distribution of constraint recoveries per epoch.}
    \label{fig:counts_boxplot_per_epoch}
\end{figure}

Figure~\ref{fig:counts_boxplot_per_epoch} shows that the guided agent maintains a consistently high compliance rate, recovering from most initial violations within the same turn via deterministic fallback generation.  
On average, over 60\% of invalid outputs are corrected immediately, showing that verifiable constraint checks stabilize multi-turn behavior.

We interpret \textit{reliability} as the consistency of constraint compliance across trajectories.  
The narrow variance in compliance ratios and task rewards indicates that the guided agent performs predictably and stably across runs, even as constraint complexity increases.  
Together, these results demonstrate that integrating explicit, feedback-guided constraint handling into generation leads to both \textit{verifiable} and \textit{reliable} multi-turn behavior.

\begin{figure}[htbp]
    \centering
    \includegraphics[width=0.92\columnwidth]{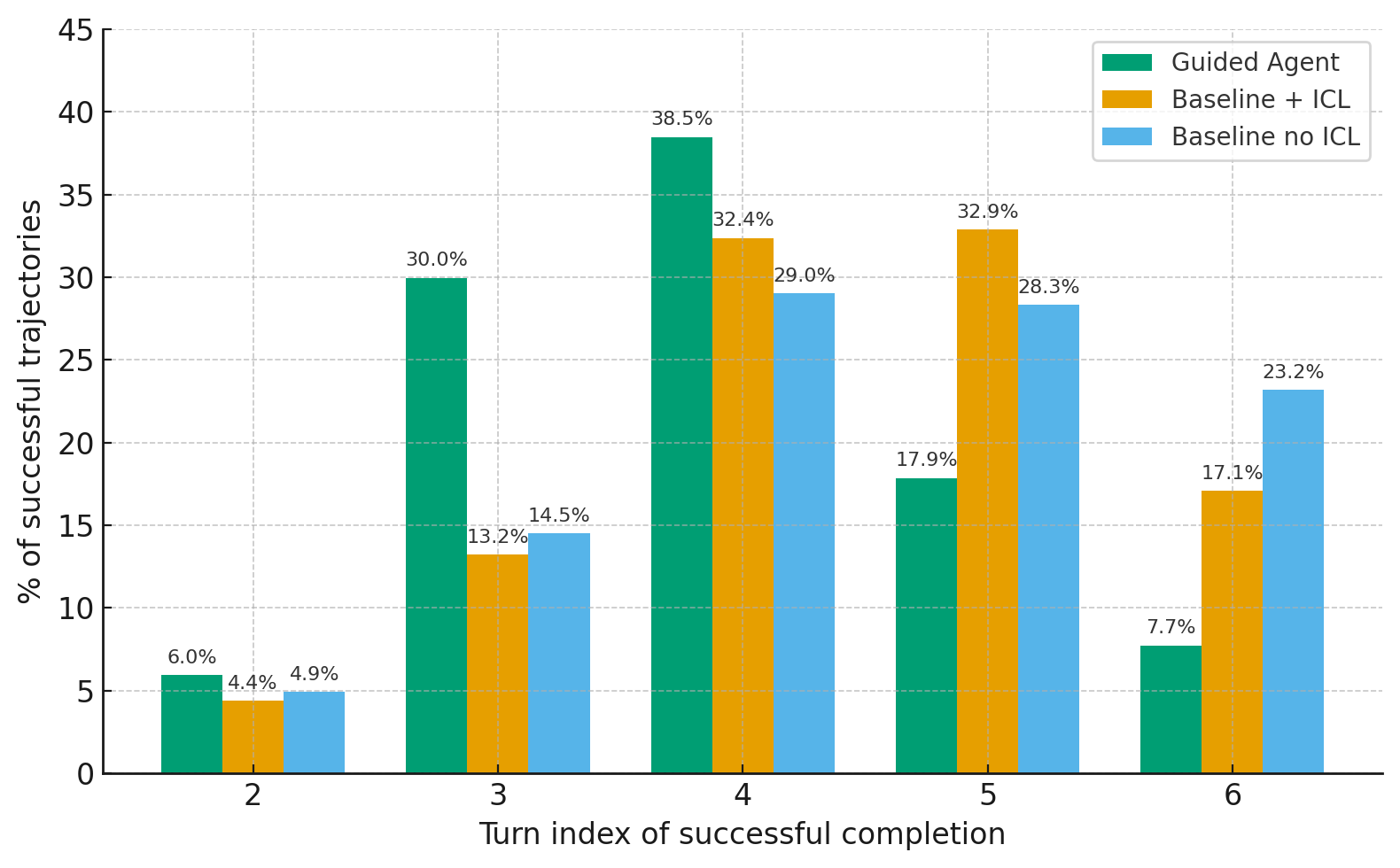}
    \caption{Wordle: \% of successful completions by turn.}
    \label{fig:wordle_success_by_turn_count}
\end{figure}

Finally, Figure~\ref{fig:wordle_success_by_turn_count} shows that the guided agent not only achieves higher success rates but also completes tasks in fewer turns, reflecting greater efficiency and behavioral stability.  
These findings confirm that our framework provides trustworthy, constraint-compliant task completion.

\section{Related Work}
\label{sec:related_work}
We classify efforts to build trustworthy multi-turn LLM agents into two directions: enhancing \textit{capability and autonomy} and improving \textit{reliability and verification}.
The first focuses on how models reason, plan, and act coherently across turns.
Frameworks such as ReAct~\cite{yao2023react}, Reflexion~\cite{shinn2023reflexion}, Toolformer~\cite{schick2023toolformer},
AutoGPT~\cite{autogpt2023}, BabyAGI~\cite{babyagi2023}, MemGPT~\cite{xu2024memgpt}, and
Voyager~\cite{wang2023voyager} extend autonomy via integrated reasoning, memory, and tool use.
However, they treat \textit{trust as an implicit outcome of competence}: agents are deemed reliable if they appear successful.
Their control dynamics remain embedded within opaque text-generation loops, lacking mechanisms to verify whether internal reasoning aligns with task constraints or feedback.
They perform well on short-horizon reasoning but struggle with cumulative, verifiable behavior over extended interactions.

Recent work makes reasoning more explicit through structured planning and cognitive control.
Frameworks such as Tree of Thoughts~\cite{yao2024tree}, Graph of Thoughts~\cite{besta2024graph},
PlanBench~\cite{zhou2024planbench}, AgentBench~\cite{li2023agentbench}, and the Hierarchical Reasoning Model (HRM)~\cite{wang2025hrm}
organize reasoning into explicit or multi-level structures.
These approaches advance transparency but still depend on prompt-level orchestration or architectural hierarchy rather than persistent modules that can learn and refine reasoning strategies.
Even when reasoning is externalized, uncertainty and explanation faithfulness remain open challenges~\cite{ganguly2025grammars, matton2025walktalk}.

The second direction targets \textit{verification and constraint satisfaction}.
SelfCheckGPT~\cite{manakul2023selfcheckgpt} verify or constrain outputs during or after decoding;
faithful reasoning and process supervision~\cite{lightman2023let} improve interpretability.
While effective for factuality and safety, these methods regulate behavior only \textit{post hoc} and remain external to the agent’s learning process.
Recent studies reveal similar reliability gaps: LLM judges show inconsistent validation under task indeterminacy~\cite{guerdan2025llmjudge}, and instruction-following models misestimate their own uncertainty~\cite{apple2025uncertainty}.
Efforts to improve evaluation diversity and consistency, especially in structured domains~\cite{zhou2025reliableeval}, further underscore the need for integrated, environment-aware verification.

Our framework unifies these directions by embedding capability, planning, and verifiability within a single feedback-driven system.
It aligns with broader efforts to formalize trust—through certified guarantees~\cite{li2025certifiedtrustworthiness}, human-centered trust calibration~\cite{swoopes2025stochasticitytrust}, and conceptual mappings of trustworthiness dimensions~\cite{decerqueira2025mappingtrustworthiness}—but differs by operationalizing trust as behavioral guidance within the agent’s own feedback loop.

\section{Conclusion}
\label{sec:conclusion}

We address the challenge of building trustworthy multi-turn LLM agents capable of reliable and verifiable task completion.  
We introduce a framework for behavioral guidance that embeds LLMs in an action--observation--reward loop and augments them with adaptive modules for task profiling, structured reasoning, and constraint-compliant generation.  

Our initial experiments show that guided agents achieve higher task success, more consistent reasoning, and stronger constraint compliance than baseline models.  
While our findings represent an early step, they suggest that structured behavioral guidance can make LLM behavior more reliable and trustworthy.

\bibliography{aaai2026}

\end{document}